\documentclass[final]{cvpr}
\usepackage{times}
\usepackage{epsfig}
\usepackage{graphicx}
\usepackage{amsmath}
\usepackage{amssymb}

\usepackage{multirow}
\usepackage{caption}
\usepackage{subcaption}
\usepackage{booktabs}
\usepackage{enumitem}

\usepackage[pagebackref=true,breaklinks=true,colorlinks,bookmarks=false]{hyperref}

\def\cvprPaperID{4492} 
\def\confYear{CVPR 2021}

\begin{document}
	
	\title{Refine Myself by Teaching Myself : \\ Feature Refinement via Self-Knowledge Distillation}
	\author{Mingi Ji$^{1}$ \hspace{0.1cm} Seungjae Shin$^{1}$ Seunghyun Hwang$^{2}$ \hspace{0.1cm} Gibeom Park$^{1}$ \hspace{0.1cm} Il-Chul Moon$^{1,3}$ \\
		\small $^1$Korea Advanced Institute of Science and Technology (KAIST) \hspace{0.1cm} $^2$Looko Inc.\hspace{0.1cm} $^3$Summary.AI\\
		{\tt\scriptsize qwertgfdcvb@kaist.ac.kr tmdwo0910@kaist.ac.kr shhwang@acloset.app pgb1227@kaist.ac.kr icmoon@kaist.ac.kr}
	}
	
	\maketitle
	\thispagestyle{empty}
	\begin{abstract}
		\textit{Knowledge distillation} is a method of transferring the knowledge from a pretrained complex teacher model to a student model, so a smaller network can replace a large teacher network at the deployment stage. To reduce the necessity of training a large teacher model, the recent literatures introduced a self-knowledge distillation, which trains a student network progressively to distill its own knowledge without a pretrained teacher network. While Self-knowledge distillation is largely divided into a data augmentation based approach and an auxiliary network based approach, the data augmentation approach looses its local information in the augmentation process, which hinders its applicability to diverse vision tasks, such as semantic segmentation. Moreover, these knowledge distillation approaches do not receive the refined feature maps, which are prevalent in the object detection and semantic segmentation community. This paper proposes a novel self-knowledge distillation method, Feature Refinement via Self-Knowledge Distillation (FRSKD), which utilizes an auxiliary self-teacher network to transfer a refined knowledge for the classifier network. Our proposed method, FRSKD, can utilize both soft label and feature-map distillations for the self-knowledge distillation. Therefore, FRSKD can be applied to classification, and semantic segmentation, which emphasize preserving the local information. We demonstrate the effectiveness of FRSKD by enumerating its performance improvements in diverse tasks and benchmark datasets. The implemented code is available at https://github.com/MingiJi/FRSKD. 		
	\end{abstract}
	
	\section{Introduction}
	\begin{figure}[t]
		\centering
		\includegraphics[width=\columnwidth]{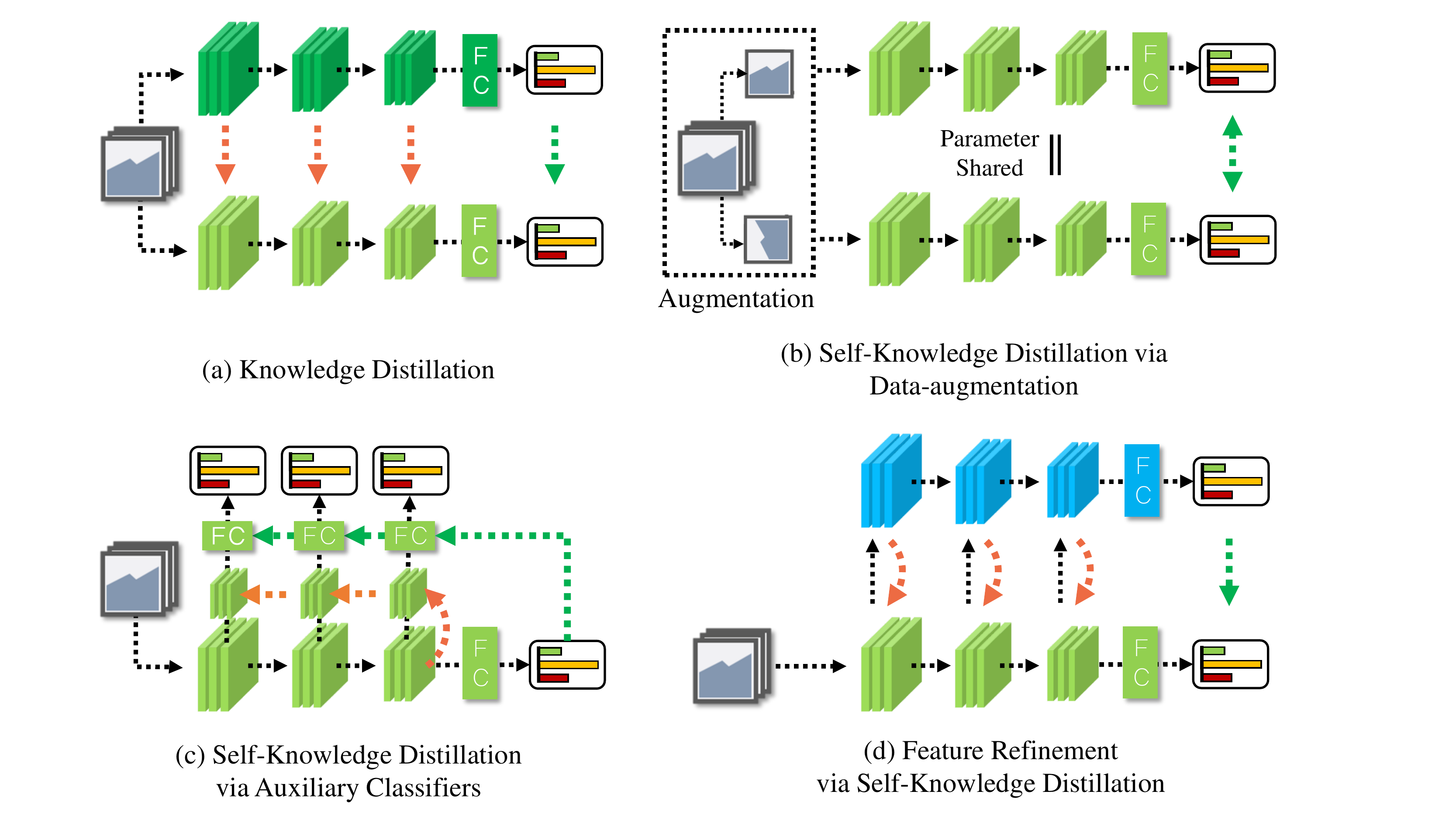}
		\caption{Comparison of various distillation methods. The black line is the forward path; the green line is the soft label distillation; and the orange line is the feature distillation. (a) Conventional knowledge distillation method with pretrained teacher~\cite{hkd,fitnets,atts,ft,vid}. (b) Self-knowledge distillation method via data augmentation~\cite{ddgsd,cskd,sla}. (c) Auxiliary weak classifier based self-knowledge distillation, which creates a set of layer-wise classifiers to generate a backpropagation signal at each layer, and the layer-wise classifier produce its estimation from the layer's feature distillations of the orange line and the logit distillations of the green line~\cite{byot}. (d) Our proposed method. The original classifier provides original feature as an input to the auxiliary self-teacher network (blue blocks). Afterwards, the self-teacher network distills the refined feature-map to the original classifier (orange lines).}
		\label{fig:intro}
		\vspace{-0.5em}
	\end{figure}

	\noindent Deep neural networks (DNNs) have been applied to various fields of computer vision due to the exponential advancement of convolutional neural networks~\cite{resnet,effnet,densenet}. To distribute the success at the mobile devices, the vision task needs to overcome the limited computing resources~\cite{mobilenet,shufflenet}. To solve this problem, the model compression has been a crucial research task, and knowledge distillation has been a prominent technology with a good compression and equivalent performances~\cite{hkd}.
	
	\textit{Knowledge distillation} is a method of transferring the knowledge from a pretrained teacher network to a student network, so a smaller network can replace a large teacher network at the deployment stage. Knowledge distillation utilizes the teacher's knowledge through receiving either 1) class predictions as soft labels~\cite{hkd}; 2) penultimate layer outputs~\cite{crd,irg,rkds}, or 3) feature-maps including spatial information at the intermediate layer~\cite{fitnets,afd,overhaul}. Whereas the knowledge distillation enables utilizing the larger network in a condensed manner, the inference on such large network, a.k.a. the teacher network, becomes an ultimate burden of its practical usages. In addition, pretraining the large network requires substantial computational resources to prepare the teacher network.
	
	To reduce such necessity of training a large network, the recent literatures introduce an alternative knowledge distillation~\cite{ban,dml}; which is a distillation from a pretrained network with the same architecture of the student network. This knowledge distillation is still known to be informative to the student network with the same scale. Moreover, there are literatures on a self-knowledge distillation, which trains the student network progressively to distill and to regularize its own knowledge without the pretrained teacher network~\cite{one,cskd,sla,ddgsd,byot}. The self-knowledge distillation is different from the previous knowledge distillation because it does not require a prior preparation of the teacher network.
	
	Self-knowledge distillation is largely divided into a data augmentation based approach and an auxiliary network based approach. The data augmentation based self-knowledge distillation induces a consistent prediction of relevant data, i.e. the different distorted versions of a single instance or a pair of instances from the same class~\cite{ddgsd,cskd,sla}. The auxiliary network based approach utilizes additional branches in the middle of the classifier network, and the additional branches are induced to make similar outputs via knowledge transfer~\cite{one,byot}. However, these approaches depend on the auxiliary network, which has the same or less complexity than classifier network; so it is hard to generate a refined knowledge, either by features, which are the output of the convolutional layers, or soft labels, for the classifier network~\cite{ddgsd,cskd,sla,one,byot}. Furthermore, the data augmentation based approaches are vulnerable to lose the local information between instances, such as differently distorted instances or rotated instances. Therefore, it is hard to utilize the feature distillation, which is well known technique for improving the performances in general knowledge distillations~\cite{ddgsd,cskd,sla}.
	
	To deal with the limitation of existing self-knowledge distillation, this paper proposes a novel self-knowledge distillation method, Feature Refinement via Self-Knowledge Distillation (FRSKD), which introduces an auxiliary self-teacher network to enable the transfer of a refined knowledge to the classifier network. Figure~\ref{fig:intro} shows the difference between FRSKD and existing knowledge distillation methods. Our proposed method, FRSKD, can utilize both soft label and feature-map distillations for the self-knowledge distillation.
	
	Therefore, FRSKD can be applied to classification and semantic segmentation, which emphasize preservation of the local information. FRSKD shows the state-of-the-art performances in image classification task on various datasets, compared to other self-knowledge distillation methods. In addition, FRSKD improves the performances on semantic segmentation. Besides, FRSKD is compatible with other self-knowledge distillation methods as well as data augmentation. We demonstrate the compatibility of FRSKD with large performance improvements through various experiments.
	
	
	\section{Related Work}
	\noindent\textbf{Knowledge distillation~~}
	The goal of \textit{knowledge distillation} is to effectively train a simpler network, a.k.a. a student network, by transferring the knowledge of a pretrained complex network, a.k.a. teacher network. Here, knowledge includes the features at the hidden layers or the logits at the final layer, etc. Hinton et al. proposed a method of knowledge distillation by transferring the teacher network's output logit to the student network~\cite{hkd}. The intermediate layer distillation methods were then introduced, so such methods utilize the teacher network's knowledge from either the convolutional layer with feature-map level preserving localities~\cite{fitnets,atts,ft,fsp,lit}; or penultimate layer~\cite{cckd,crd,irg,rkds,sp}. For the feature-map distillations, the prior works induced the student to imitate 1) feature of the teacher network~\cite{fitnets}, 2) abstracted attention map of the teacher network~\cite{atts}, or 3) FSP matrix of the teacher network~\cite{fsp}. For the penultimate layer distillations, the existing literature utilized the relation between instances as knowledge, i.e. the cosine similarity between feature sets at the same penultimate layer from pair of instances~\cite{cckd,crd,irg,rkds,sp}. In addition, prior experiments concluded that these distillation methods on different layers perform better when different distillations are jointly applied. However, there are two distinct limitations: 1) knowledge distillation requires pretraining of the complex teacher model, and 2) variation of the teacher networks will result in different performances with the same student network.
	
	\noindent\textbf{Self-knowledge distillation~~} 
	\textit{Self-knowledge distillation} enhances the effectiveness of training a student network by utilizing its own knowledge without a teacher network. First, some approaches utilize an auxiliary network for self-knowledge distillation. For example, BYOT introduced a set of auxiliary weak classifier networks that classify the output with the features from the middle hidden layers~\cite{byot}. The weak classifier networks of BYOT are trained by the joint supervision of estimated logit values and true supervision. ONE utilized additional branches to diversify model parameters and estimated features at the middle layer. This diversification is aggregated through the ensemble method, and the ensembled output creates the joint back-propagation signal shared by the branches~\cite{one}. The commonality of these auxiliary network approaches is the utilization of Adhoc structure at the same level or weak classifier network at the middle layer. Hence, these approaches without enhanced network may suffer from a lack of more refined knowledge.
	
	
	Secondly, data augmentation was used for self-knowledge distillation as well. DDGSD induced the consistent prediction by providing differently augmented instances, so the classifier network would face variations of instances~\cite{ddgsd}. CSKD used logits of other instances belonging to the same class for regularization purposes, so the classifier network would predict similar outcomes for the same classes~\cite{cskd}. However, the data augmentations do not necessarily preserve the spatial information, i.e. simple flipping would ruin the feature locality, so the feature-map distillation is difficult to be applied in the line of data augmentations. SLA proposed augmenting the data label by combining the self-supervision task with the original classification task. The self-supervision takes augmentations such as input rotations, and the ensembled output of the augmented instances provides additional supervision for the backpropagation~\cite{sla}. 
	
	Ideally, the feature-map distillation can be improved by extracting the refined knowledge from a complex model. However, the auxiliary structure of the prior works did not provide such a method to make feature more complex. On the contrary, the data augmentation may increase data variations or refinements in some cases, but their variability can hinder the feature-map distillation because the variation will prevent consistent locality modelling from the parameter side. Therefore, we conjecture that supplying the refinements to the feature-map distillation can be a breakthrough in the auxiliary network-based self-knowledge distillation. Hence, we suggest the auxiliary self-teacher network to generate a refined feature-map as well as its soft label. To our knowledge, this paper is the first work on a self-teacher network to generate a refined feature-maps from a single instance.
	
	\noindent\textbf{Feature networks~~}
	Our suggested structure of the auxiliary self-teacher network is developed from the feature networks used in the object detection field. The aggregation of feature with various scales is one of key for processing multi-scaled features, and the object detection field has investigated this scale variations in ~\cite{fpn,fpn2,fpn3,panet,bifpn}. Our auxiliary self-teacher network generates a refined feature-map by adapting the network that deals with multi scaled features to the knowledge distillation purpose. While we will explain the adaptation of this structure in Section~\ref{sec:self-teacher}, this subsection enumerates the recent developments of the feature networks. FPN utilized a top-down network to simultaneously exploit 1) abstract information from the upper layers and 2) information on small objects at the lower layers of the backbone network~\cite{fpn}. PANet introduced additional bottom-up network for FPN to enable a short path connection between a layer for detection and a layer of backbone~\cite{panet}. BiFPN proposed a more efficient network structure, using top-down and bottom-up networks as same as PANet~\cite{bifpn}.
	This paper proposes an auxiliary self-teacher network, which is altered from the structure of BiFPN to be appropriate for the classification task. In addition, the feature-map distillation has become efficient by the alteration of the network structure because the self-teacher network requires fewer computations than BiFPN by varying the channel dimension according to the depth of the feature-maps.
	
	\begin{figure*}[t]
		\centering
		\includegraphics[width=2\columnwidth]{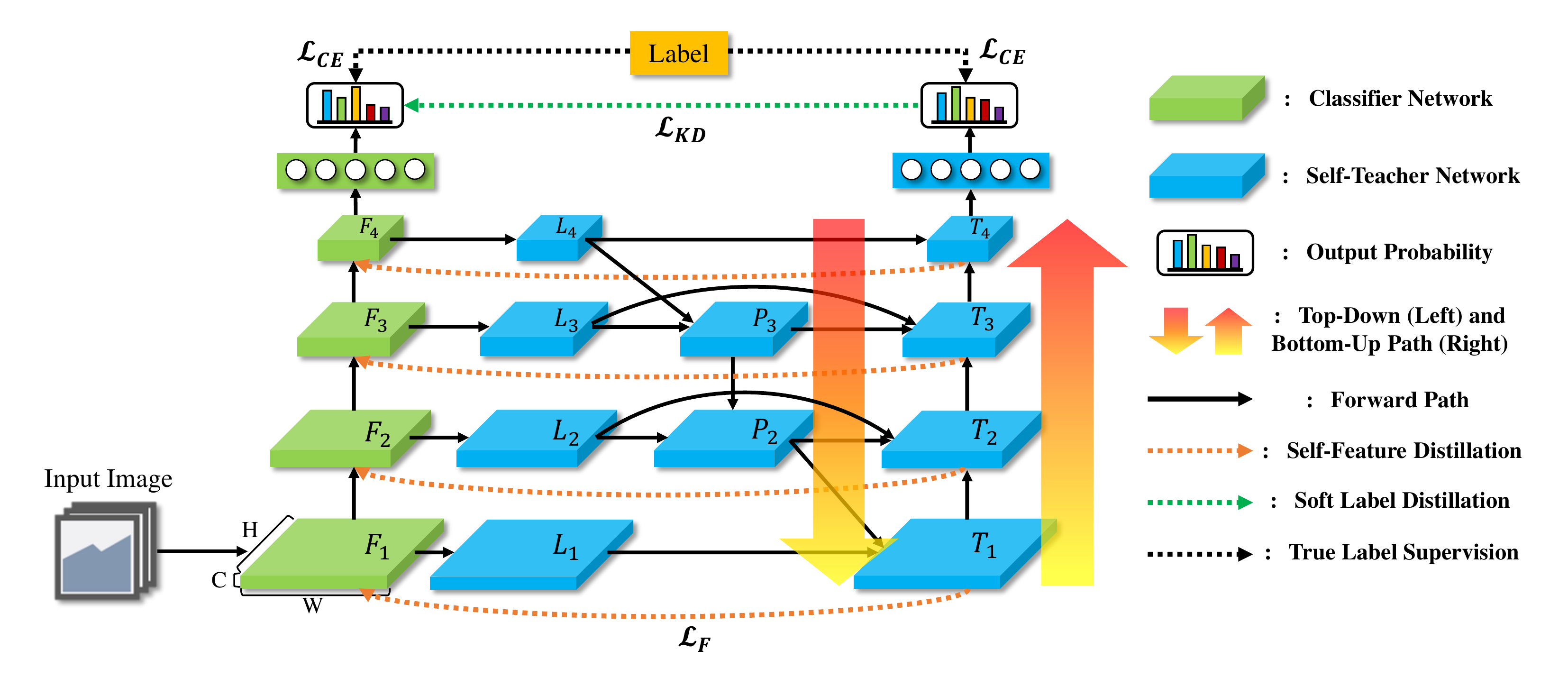}
		\caption{Overview of our proposed self-knowledge distillation method, Feature Refinement via Self-Knowledge Distillation (FRSKD). The top-down path and the bottom-up path aggregates different sized features and provide the refined feature-map to the original classifier network. Exploiting feature-map of the self-teacher network, FRSKD performs distillation on refined feature-map and soft label.}
		\label{fig:method}
	\end{figure*}
	
	\section{Method}
	\noindent This section introduces a feature refinement self-knowledge distillation (FRSKD). Figure \ref{fig:method} shows the overview of our distillation method, which is further discussed in Section~\ref{sec:self-teacher} from the perspective of the self-teacher network. Then, we review the training procedure of our self-knowledge distillation in Section \ref{sec:distill}.
	
	\noindent\textbf{Notations~~}
	Let $D=\{(\boldsymbol{x}_1,y_1), (\boldsymbol{x}_2,y_2)), ..., (\boldsymbol{x}_N,y_N))\}$ be a set of labeled instances where $N$ is its size; let $F_{i,j}$ be a feature map of the $j$-th block of the classifier network for the $i$-th sample; and let $c_j$ be a channel dimension of the $j$-th block of the classifier network. For notation simplicity, we omit the index $i$ in the rest of this paper.
	
	\subsection{Self-Teacher Network} \label{sec:self-teacher}
	The main purpose of the self-teacher network is providing a refined feature-map and its soft label for the classifier network to itself. The inputs of the self-teacher network are the feature-maps of the classifier network, $F_1, ..., F_n$, which assumes the $n$ blocks of the classifier network. We model the self-teacher network by modifying the structure of BiFPN for the classification task. Specifically, we adapt the top-down path and the bottom-up path from PANet and BiFPN~\cite{bifpn,panet}. Before the top-down path, we utilize the lateral convolutional layers as following:
	\begin{equation}
		L_i = Conv(F_i; d_i) 
		\label{eq:lateral}
	\end{equation}
	$Conv$ is a convolutional operation with an output dimension of $d_i$. Unlike the existing lateral convolutional layers with $d_i$ dimensions that is fixed at the network setup, we design $d_i$ that depends upon the channel dimension of the feature map, $c_i$. We set $d_i$ as $d_i = w \times c_i$ with a channel width parameter, $w$. For the classification task, it is natural to set a higher channel dimension for the deeper layers. Therefore, we adjust the channel dimension of each layer to contain information of its feature-map depth. Also, this design reduces the computation of the lateral layers. 
	
	The top-down path and the bottom-up path aggregate different features as the below:
	\begin{align}
		& P_i = Conv(w^P_{i,1} \cdot L_i + w^P_{i,2} \cdot Resize(P_{i+1}); d_i)   \\
		& T_i = Conv(w^T_{i,1} \cdot L_i + w^T_{i,2} \cdot P_i + w^T_{i,3} \cdot Resize(T_{i-1}); d_i) \nonumber
	\end{align} \label{eq:path}
	$P_i$ represents the $i$-th layer of the top-down path; and $T_i$ is the $i$-th layer of the bottom-up path. Similar to BiFPN~\cite{bifpn}, the forward pass connection has a different structure depending on the depth of the layers. In Figure~\ref{fig:method}, in the case of the shallowest bottom-up path layer $T_1$ and the deepest bottom-up path layer $T_4$, each directly utilizes the lateral layer $L_1$ and $L_4$ as inputs respectively for efficiency, instead of using the features of the top-down path. In these settings, to create a top-down structure that connects all the shallowest layer, middle layer and deepest layer, the two diagonal connections for forward propagations are added: 1) the connection from the last lateral layer, $L_4$, to the penultimate layer of the top-down path, $P_3$; and 2) the connection from the $P_2$, to the first layer of the bottom-up path, $T_1$.
	
	We apply a fast normalized fusion with parameters, such as $w^P$ and $w^T$ \cite{bifpn}. We use a bilinear interpolation for the up-sampling, and we use the max-pooling for the down-sampling, as $Resize$ operator. For efficient calculations, we use a depth-wise convolution for convolutional operations \cite{mobilenet}. We conduct various experiments, and we analyze the results, according to the self-teacher network structure in Section~\ref{sec:analysis}. Finally, we attach the fully connected layer on the top of the bottom-up path to predict the output class, and the self-teacher network provides its soft label, $\hat{p}_t=\text{softmax}(f_t(\boldsymbol{x};\theta_t))$ where $f_t$ denotes the self-teacher network, parameterized by $\theta_t$.
	
	\subsection{Self-Feature Distillation} \label{sec:distill}
	Our proposed model, FRSKD, utilizes the output of the self-teacher network, the refined feature-map, $T_i$, and the soft label, $\hat{p}_t$. Firstly, we add the feature distillation, which induces the classifier network to mimic the refined feature-map. For feature distillations, we adapt the attention transfer \cite{atts}. Equation~\ref{eq:att} defines the feature distillation loss, $\mathcal{L}_F$:
	\begin{equation}
		\mathcal{L}_F(T,F;\theta_c,\theta_t)=\Sigma_{i=1}^n||\phi(T_i)-\phi(F_i)||_2
		\label{eq:att}
	\end{equation}
	where $\phi$ is a combination of channel-wise pooling function with $L_2$ normalization \cite{atts}; and $\theta_c$ is parameter of the classifier network. $\phi$ abstracts the spatial information of the feature-map. Therefore, $\mathcal{L}_F$ makes the classifier network learn the locality of the refined feature-map from the self-teacher network. In addition, it is possible to train a classifier network to mimic the refined feature-map exactly~\cite{fitnets,overhaul}; or utilize the transformation of the feature-map~\cite{fsp}. Unless noted, this paper utilizes the attention transfer-based feature distillation, and we discuss the methodology of the feature distillation in Section~\ref{sec:analysis}.
	
	Similar to other self-knowledge distillation methods, FRSKD also performs the distillation through the soft label, $\hat{p}_t$, as following: 
	\begin{align}
		&    \mathcal{L}_{KD}(\boldsymbol{x};\theta_c,\theta_t,K) \\ \nonumber & = D_{KL}(\text{softmax}(\frac{f_c(\boldsymbol{x};\theta_c)}{K})||\text{softmax}(\frac{f_t(\boldsymbol{x};\theta_t)}{K})
	\end{align}
	where $f_c$ is the classifier network; and $K$ is the temperature scaling parameter. Also, the classifier network and the self-teacher network learn a ground-truth label using a cross-entropy loss, $\mathcal{L}_{CE}$. By integrating the loss functions above, we construct the following optimization objective:
	\begin{align}
		&\mathcal{L}_{FRSKD}(\boldsymbol{x},y;\theta_c,\theta_t,K) \\ & \nonumber =\mathcal{L}_{CE}(\boldsymbol{x},y;\theta_c)+\mathcal{L}_{CE}(\boldsymbol{x},y;\theta_t) \\ \nonumber
		& ~~~~ + \alpha \cdot \mathcal{L}_{KD}(\boldsymbol{x};\theta_c,\theta_t,K) + \beta \cdot \mathcal{L}_{F}(T,F;\theta_c,\theta_t)
	\end{align}
	where $\alpha$ and $\beta$ are hyperparameters; and we choose $\alpha\in[1,2,3]$ and $\beta\in[100,200]$; further details are explained in Appendix \textit{Sensitivity Analysis}. Optimization is initiated by backpropagations at the same time for both classifier and self-teacher networks. To prevent the model collapse issue~\cite{vat}, FRSKD updates the parameters by the distillation loss, $\mathcal{L}_{KD}$ and $\mathcal{L}_F$, which is only applied to the student network.
	
	\section{Experiments}
	We evaluate our self-knowledge distillation method on various tasks: classification and semantic segmentation. Throughout this section, we mainly use three settings; utilizing the distillation of soft label only (FRSKD\textbackslash F); optimizing by $\mathcal{L}_{FRSKD}$ with the distillation of refined feature-map and its soft label (FRSKD); and attaching the data augmentation based on the self-knowledge distillation, SLA-SD, with our method (FRSKD+SLA)~\cite{sla}.
	
	\subsection{Classification}
	\noindent\textbf{Datasets~~}
	We demonstrate the effectiveness of FRSKD on various classification datasets: CIFAR-100~\cite{cifar100}, TinyImageNet, Caltech-UCSD Bird (CUB200)~\cite{cub}, MIT Indoor Scene Recognition (MIT67)~\cite{mit67}, Stanford 40 Actions (Stanford40)~\cite{stan40}, Stanford Dogs (Dogs)~\cite{stan_dog}, and ImageNet~\cite{imagenet}. Cifar-100 and TinyImageNet consist of small scaled images, and we resized TinyImageNet images to meet the same size of CIFAR-100 (32$\times$32). CUB200, MIT67, Stanford40 and Dogs are datasets for the fine-grained visual recognition (FGVR) tasks. FGVR contains fewer data instances per class compared to CIFAR-100 and ImageNet. ImageNet is a large scaled dataset to validate our method to practically test the model.
	
	\noindent\textbf{Implementation details~~} 
	We use ResNet18 and WRN-16-2 \cite{resnet,wrn} for CIFAR-100 and TinyImageNet. To adapt ResNet18 to small sized datasets, we modify the first convolution layer of ResNet18 as a kernel size of $3\times3$, a single stride, and a single padding. We remove a max-pooling operation, as well. We use the standard ResNet18 for FGVR tasks; and we apply both ResNet18 and ResNet34 to ImageNet.
	
	For all classification experiments, we use stochastic gradient descents (SGD) with an initial learning rate of 0.1 and weight decay as 0.0001. We set total epochs as 200, and we divide the learning rate by 10 at epoch 100 and 150 for CIFAR-100, TinyImageNet and FGVR. For ImageNet, we set total epochs as 90, and we divide the learning rate by 10 at epoch 30 and 60. We set the batch size as 128 for CIFAR-100 and TinyImageNet; 32 for FGVR; and 256 for ImageNet. We use standard data augmentation methods for all experiments, i.e. random cropping and flipping. In terms of hyperparameters, we set $\alpha$ as two and $\beta$ as 100 for CIFAR-100; $\alpha$ as three and $\beta$ as 100 for TinyImageNet; and $\alpha$ as one and $\beta$ as 200 for FGVR and ImageNet. Additionally, we set the temperature scaling parameter, $K$, as four; and we set the channel width parameter, $w$, as two for all experiments. Further details are enumerated in Appendix \textit{Implementation Details}.
	
	\noindent\textbf{Baselines~~}
	We compare FRSKD to a standard classifier (named as Baseline), which doesn't utilize the distillation, with cross entropy based loss and six self-knowledge distillation methods, which make a total of seven baselines.
	\begin{itemize}
		\item \textbf{ONE}~\cite{one} exploits an ensembled prediction of additional branches as the soft label.
		\item \textbf{DDGSD}~\cite{ddgsd} generates different distorted versions of a single instance, and DDGSD trains to produce consistent prediction for the distorted data. 
		\item \textbf{BYOT}~\cite{byot} applies auxiliary classifiers utilizing the outputs of intermediate layers, and BYOT trains auxiliary classifiers by ground-truth labels and signals from network itself, such as the predicted logit or the feature-map.
		\item \textbf{SAD}~\cite{sad} focuses on a lane detection by a layer-wise attention distillation in network itself.
		\item \textbf{CS-KD}~\cite{cskd} forces a consistent prediction for the same class by utilizing the prediction of other instances within the same class as the soft label.
		\item \textbf{SLA-SD}~\cite{sla} trains a network with an original classification task and a self-supervised task jointly by utilizing the label augmentation. SLA-SD exploits an aggregated prediction as the soft label.
	\end{itemize}
	We utilize the available official code for implementation \cite{cskd,sla,one}. Otherwise, we implement models according to the corresponding papers. We apply same training setting according to the dataset, and we tune the hyperparameters of baseline models.
	
	\begin{table}[h]
		\centering
		\resizebox{\columnwidth}{!}{%
			\begin{tabular}{@{}ccccccc@{}}
				\toprule
				\multirow{2}{*}{Methods} & & \multicolumn{2}{c}{CIFAR100} & & \multicolumn{2}{c}{TinyImageNet} \\ \cmidrule(lr){3-4} \cmidrule(l){6-7} 
				& & \multicolumn{1}{c}{WRN-16-2} & \multicolumn{1}{c}{ResNet18}  &  &\multicolumn{1}{c}{WRN-16-2}  &   \multicolumn{1}{c}{ResNet18}    \\ \midrule
				Baseline                 & & 70.42$\pm$\scriptsize{0.08}&73.80$\pm$\scriptsize{0.60}& &51.05$\pm$\scriptsize{0.20}&54.60$\pm$\scriptsize{0.33}\\
				ONE                      & & 73.01$\pm$\scriptsize{0.23}&76.67$\pm$\scriptsize{0.66}& &52.10$\pm$\scriptsize{0.20}&57.53$\pm$\scriptsize{0.39}\\
				DDGSD                    & & 71.96$\pm$\scriptsize{0.05}&76.61$\pm$\scriptsize{0.47}& &51.07$\pm$\scriptsize{0.24}&56.46$\pm$\scriptsize{0.24}\\
				BYOT                     & & 70.22$\pm$\scriptsize{0.26}&76.68$\pm$\scriptsize{0.07}& &50.33$\pm$\scriptsize{0.03}&56.61$\pm$\scriptsize{0.30}\\
				SAD                      & & 70.31$\pm$\scriptsize{0.45}&74.65$\pm$\scriptsize{0.33}& &51.26$\pm$\scriptsize{0.39}&54.45$\pm$\scriptsize{0.06}\\
				CS-KD                    & & 71.79$\pm$\scriptsize{0.68}&77.19$\pm$\scriptsize{0.05}& &50.08$\pm$\scriptsize{0.18}&56.46$\pm$\scriptsize{0.10}\\
				SLA-SD                   & & 73.00$\pm$\scriptsize{0.45}&77.52$\pm$\scriptsize{0.30}& &50.77$\pm$\scriptsize{0.33}&58.48$\pm$\scriptsize{0.44}\\ \midrule
				FRSKD\textbackslash F              & & 73.12$\pm$\scriptsize{0.06}&77.64$\pm$\scriptsize{0.12}&  & \underline{52.91$\pm$\scriptsize{0.30}} &59.50$\pm$\scriptsize{0.15}\\
				FRSKD          & & \underline{73.27$\pm$\scriptsize{0.45}}&\underline{77.71$\pm$\scriptsize{0.14}}& & \textbf{53.08$\pm$\scriptsize{0.33}}&\underline{59.61$\pm$\scriptsize{0.31}}\\
				FRSKD+SLA    & &  \textbf{75.43$\pm$\scriptsize{0.21}}  & \textbf{82.04$\pm$\scriptsize{0.16}}  &   &   51.83$\pm$\scriptsize{0.37}     &  \textbf{63.58$\pm$\scriptsize{0.04}}              \\ \bottomrule
			\end{tabular}
		}
		\caption{Performance comparison on CIFAR-100 and TinyImageNet. Experiments are repeated three times, and we report average and standard deviation of the accuracy of the last epoch. The best performing model is indicated as boldface. The second-best model is indicated as underline.}
		\label{table:cifar}
	\end{table}
	
	\noindent\textbf{Performance comparison~~}
	Table~\ref{table:cifar} shows the classification accuracy on CIFAR-100 and TinyImageNet with two different classifier network structures. Most of the self-knowledge distillation methods improve the performance of the standard classifier. Compared to baselines, FRSKD consistently shows better performance than other self-knowledge distillation methods. Furthermore, FRSKD\textbackslash F, which does not utilize the feature distillation, shows better performance than other baselines. This result shows that the soft label of the self-teacher network outperforms the data augmentation based methods. Additionally, the effect of the feature distillation is demonstrated by the outperformance of FRSKD compared to FRSKD\textbackslash F. Our proposed model, FRSKD, is not dependent on the data augmentation, so FRSKD is compatible with other data augmentation based self-knowledge distillation methods, such as SLA-SD. Hence, we conduct experiments by integrating FRSKD and SLA-SD (FRSKD+SLA), and FRSKD+SLA shows performance improvements with large margins on most experiments.
	
	\begin{table}[h]
		\centering
		\resizebox{\columnwidth}{!}{%
			\begin{tabular}{@{}ccccc@{}}
				\toprule
				Methods     & \multicolumn{1}{c}{CUB200} & \multicolumn{1}{c}{MIT67} & \multicolumn{1}{c}{Dogs} & \multicolumn{1}{c}{Stanford40} \\ \midrule
				Baseline    &51.72$\pm$\scriptsize{1.17}&55.00$\pm$\scriptsize{0.97}&63.38$\pm$\scriptsize{0.04}&42.97$\pm$\scriptsize{0.66}\\
				ONE         &54.71$\pm$\scriptsize{0.42}&56.77$\pm$\scriptsize{0.76}&65.39$\pm$\scriptsize{0.59}&45.35$\pm$\scriptsize{0.53}\\
				DDGSD       &58.49$\pm$\scriptsize{0.55}&59.00$\pm$\scriptsize{0.77}&69.00$\pm$\scriptsize{0.28}&45.81$\pm$\scriptsize{1.79}\\
				BYOT        &58.66$\pm$\scriptsize{0.51}&58.41$\pm$\scriptsize{0.71}&68.82$\pm$\scriptsize{0.15}&48.51$\pm$\scriptsize{1.02}\\
				SAD         &52.76$\pm$\scriptsize{0.57}&54.48$\pm$\scriptsize{1.30}&63.17$\pm$\scriptsize{0.56}&43.52$\pm$\scriptsize{0.06}\\
				CS-KD       &64.34$\pm$\scriptsize{0.08}&57.36$\pm$\scriptsize{0.37}&68.91$\pm$\scriptsize{0.40}&47.23$\pm$\scriptsize{0.22}\\
				SLA-SD      &56.17$\pm$\scriptsize{0.71}&61.57$\pm$\scriptsize{1.06}&67.30$\pm$\scriptsize{0.21}&54.07$\pm$\scriptsize{0.38}\\ \midrule
				FRSKD\textbackslash F & 62.29$\pm$\scriptsize{1.65}&61.32$\pm$\scriptsize{0.67}&69.48$\pm$\scriptsize{0.84}&53.16$\pm$\scriptsize{0.44}\\
				FRSKD & \underline{65.39$\pm$\scriptsize{0.13}}&\underline{61.74$\pm$\scriptsize{0.67}}&\underline{70.77$\pm$\scriptsize{0.20}}&\underline{56.00$\pm$\scriptsize{1.19}} \\
				FRSKD+SLA  &   \textbf{67.80$\pm$\scriptsize{1.24}} & \textbf{66.04$\pm$\scriptsize{0.31}} &  \textbf{72.48$\pm$\scriptsize{0.34}} &  \textbf{61.96$\pm$\scriptsize{0.57}}  \\ \bottomrule
			\end{tabular}
		}
		\caption{Performance comparison on FGVR. Experiments are repeated three times, and we report average and standard deviation of the accuracy of the last epoch. The best performing model is indicated as boldface. The second-best model is indicated as underline.}
		\label{table:FGVR}
	\end{table}
	
	Table~\ref{table:FGVR} shows the classification accuracy on FGVR tasks. Similar to the result of Table~\ref{table:cifar}, FRSKD shows better performance than other self-knowledge distillation methods. The superior performance of FRSKD against FRSKD\textbackslash F indicates that the effect of feature distillation is greater when using a larger image. Also, FRSKD+SLA outperforms all of the other methods with large margins, so the compatibility of FRSKD with data augmentation based self-knowledge distillation methods provides a significant advantage in FGVR tasks.
	
	\begin{table}[h]
		\centering
		\resizebox{0.6\columnwidth}{!}{%
			\begin{tabular}{cccc}
				\toprule
				Model  & Method   & Top-1                & Top-5                \\ \midrule
				\multirow{2}{*}{ResNet18}     & Baseline & 69.76                & 89.08                \\
				& FRSKD & \multicolumn{1}{c}{\textbf{70.17}} & \multicolumn{1}{c}{\textbf{89.78}} \\ \midrule
				\multirow{2}{*}{ResNet34}     & Baseline & 73.31                & 91.42                \\
				& FRSKD & \multicolumn{1}{c}{\textbf{73.75}} & \multicolumn{1}{c}{\textbf{92.11}} \\
				\bottomrule
			\end{tabular}
		}
		\caption{Performance comparison on ImageNet. The best performing model is indicated as boldface.}
		\label{table:ImageNet}
	\end{table}
	To demonstrate FRSKD on large-scaled datasets, we evaluate FRSKD on ImageNet with two backbone network alternatives, ResNet18 and ResNet34. Table~\ref{table:ImageNet} shows that FRSKD improves the performance on ImageNet.

	\subsection{Semantic Segmentation}
	We conduct an experiment on semantic segmentation to verify the efficiency of FRSKD in various domains. We follow the most of the experimental settings from \cite{overhaul}. We use a combined dataset of VOC2007 and VOC2012 \textit{trainval} as a training set; and we use the test set of VOC2007 as a validation set, which are widely used settings in semantic segmentation~\cite{voc2007,voc2012}. This experiment utilizes an EfficientDet with stacked BiFPN structure~\cite{bifpn} as a baseline. For our experiments, we stack three BiFPN layers, and we use additional two BiFPN layer as the self-teacher network. We set an initial learning rate as 0.01; we set the total epochs as 60; and we divide the learning rate by 10 at epoch 40. 
	We describe further details in Appendix \textit{Implementation Details}. Table~\ref{table:segmentation} shows that FRSKD improves the performance of the semantic segmentation models by utilizing the self-knowledge distillation from the self-teacher network.
	
	\begin{table}[h]
		\centering
		\resizebox{0.6\columnwidth}{!}{%
			\begin{tabular}{ccc}
				\toprule
				
				\multicolumn{1}{c}{Model}             & \multicolumn{1}{l}{Method} & mIOU                       \\ \midrule
				\multirow{2}{*}{EfficientDet-d0} & Baseline                   & \multicolumn{1}{c}{79.07} \\
				& FRSKD  & \textbf{80.55}                           \\ \midrule
				\multirow{2}{*}{EfficientDet-d1} & Baseline                   & \multicolumn{1}{c}{81.95}      \\
				& FRSKD                      &   \textbf{83.88}   \\ \bottomrule        
			\end{tabular}
		}
		\caption{Performance comparison on semantic segmentation task. The best performing model is indicated as boldface.}
		\label{table:segmentation}
	\end{table}
	
	\begin{figure}[h]
		\includegraphics[width=\columnwidth]{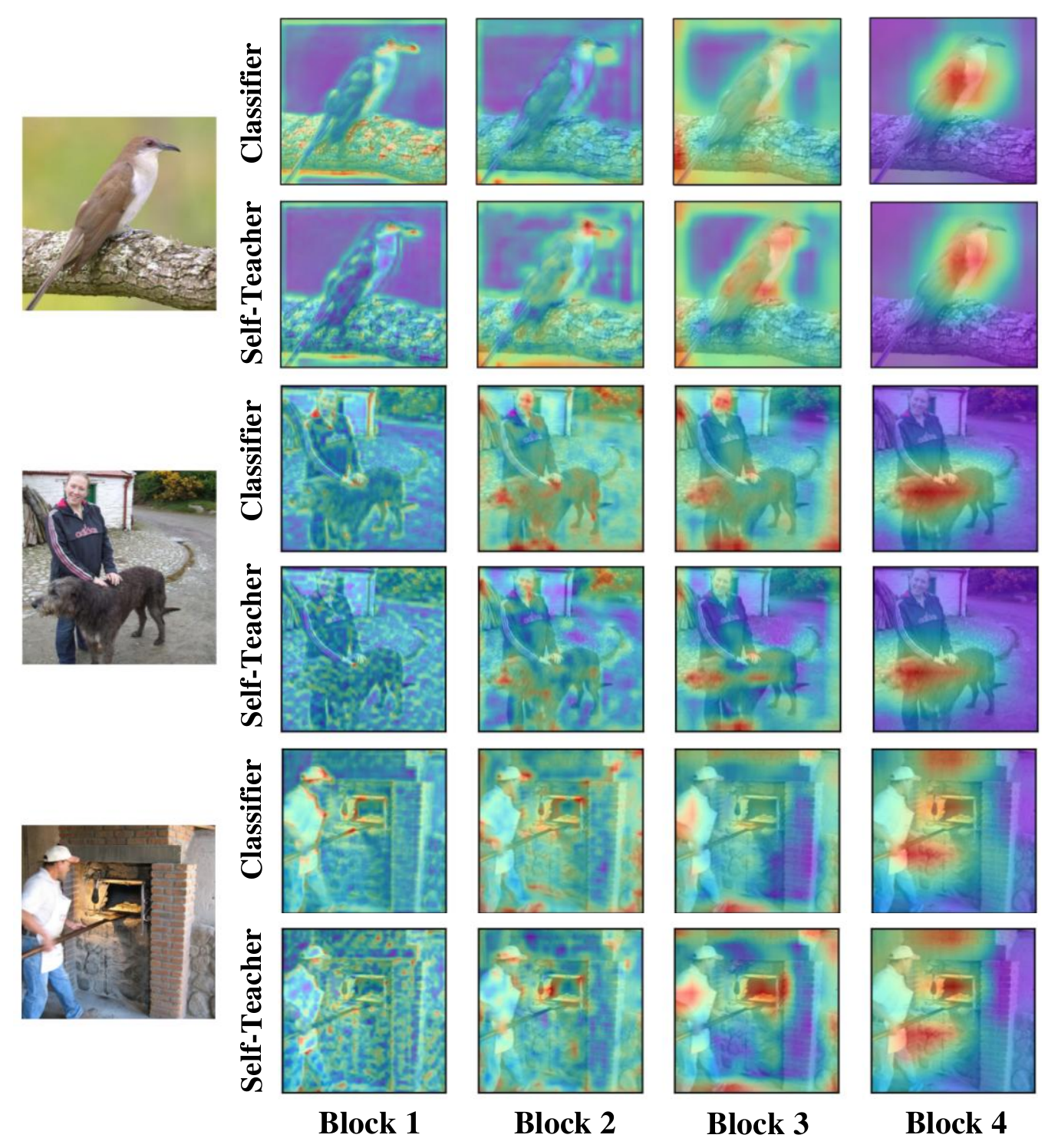}
		\caption{The block-wise attention map comparison between classifier network (first row from each data) and self-teacher network (second row from each data). From above, each data was taken from CUB200, Dogs and MIT67.}
		\label{fig:qual_block}
	\end{figure}
	
	\begin{table*}[t]
		\centering
		\begin{tabular}{ccccccccc}
			\toprule
			Method & CIFAR-100 & TinyImageNet & CUB200 & MIT67 & Dogs & Stanford40 \\ \midrule
			Baseline & 73.80$\pm$\scriptsize{0.60} & 54.60$\pm$\scriptsize{0.33} & 51.72$\pm$\scriptsize{1.17} & 55.00$\pm$\scriptsize{0.97} & 63.38$\pm$\scriptsize{0.04} & 42.97$\pm$\scriptsize{0.66}\\
			Fit+SKD & 77.03$\pm$\scriptsize{0.05} & 59.06$\pm$\scriptsize{0.20} & \underline{61.05$\pm$\scriptsize{1.05}} &  \underline{57.69$\pm$\scriptsize{0.28}} & \underline{67.50$\pm$\scriptsize{0.32}} &  \underline{51.66$\pm$\scriptsize{1.32}}\\
			OD+SKD & \underline{77.12$\pm$\scriptsize{0.09}} & \underline{59.14$\pm$\scriptsize{0.20}} & 57.44$\pm$\scriptsize{0.92} &  54.83$\pm$\scriptsize{2.63} & 66.51$\pm$\scriptsize{0.87} &  49.09$\pm$\scriptsize{0.47}\\
			FRSKD & \textbf{77.71$\pm$\scriptsize{0.14}} & \textbf{59.61$\pm$\scriptsize{0.31}} & \textbf{65.39$\pm$\scriptsize{0.13}} & \textbf{61.74$\pm$\scriptsize{0.67}} & \textbf{70.77$\pm$\scriptsize{0.20}} & \textbf{56.00$\pm$\scriptsize{1.19}}\\
			\bottomrule
		\end{tabular}
		\caption{Performance comparison according to feature distillation method of FRSKD. The feature distillation method of Fit+SKD is based on FitNet~\cite{fitnets}; OD+SKD is based on overhaul distillation~\cite{overhaul}; and FRSKD is based on attention transfer~\cite{atts}. ResNet18 is used as classifier network. The best performing model is indicated as boldface. The second-best model is indicated as underline.}
		\label{table:feature_distill}
	\end{table*}
	
	\subsection{Further Analyses on FRSKD} \label{sec:analysis}
	\noindent\textbf{Qualitative Attention map comparison}
	In order to check whether the classifier network is receiving a meaningful distillation from the self-teacher network, we conduct qualitative analyses by comparing the attention maps of each block from the classifier network and the self-teacher network. We obtain the attention map by applying a channel-wise pooling to the feature map for each block from the various datasets: CUB200, MIT67 and the Dogs dataset. We select the attention maps at the 50-th epoch to observe the distillation behaviors in the learning process.
	
	Figure~\ref{fig:qual_block} shows the differences in the block-wise attention maps for the classifier network and the self-teacher network. For the data from CUB200 dataset, which is designed to distinguish the species of bird, the cases of Block 2 and 3 illustrate that the classifier network does not capture the proper attention on the main object (bird). In contrast, the cases of Block 2 and 3 in a self-teacher network show coherent attention maps on the target object by utilizing the aggregated features. This trend can also be found in the data from the Dogs dataset. The self-teacher's attention map points to the main object (dog) relatively with a concentration compared to the attention map of the classifier that is biased toward the human, which is not the main object. The attention map comparison is also conducted on MIT67 dataset, which performs indoor scene recognition by reflecting the overall context, not the task of concentrating on a single object. In order to successfully recognize the scene class (bakery) of the data, it is important to utilize the contextual cues inside the data. From the case of block 3, unlike the classifier network, the self-teacher network pays more attention to the bread, which can be an important clue to the scene class (bakery).
	
	\begin{table}[t]
		\centering
		\resizebox{\columnwidth}{!}{%
			\begin{tabular}{ccccccccc}
				\toprule
				Type & \#channel & Parameters & FLOPs & CIFAR-100\\ \midrule
				BiFPN & 128 & $\times$ 0.30  & $\times$ 0.67  &  72.64$\pm$\scriptsize{0.12}  \\
				BiFPN & 256 & $\times$ 0.97  &  $\times$ 2.38 &  73.54$\pm$\scriptsize{0.41}  \\ 
				BiFPNc & 128 & $\times$ 0.19  &  $\times$ 0.21 & 71.70$\pm$\scriptsize{0.19}   \\ 
				BiFPNc & 256 &  $\times$ 0.59 &  $\times$ 0.68 & 73.27$\pm$\scriptsize{0.45}    \\ \bottomrule
			\end{tabular}
		}
		\caption{Performance and efficiency comparison between the self-teacher network structures. WRN-16-2 is used as classifier network on CIFAR-100. BiFPN is a structure with the same channel dimension for each layer, and BiFPNc is a structure in which the channel dimension is different depending on the depth of the layer as proposed in Section~\ref{sec:self-teacher}. \#channel is the channel dimensionl of the deepest layer, i.e. $L_n, P_n, T_n$, of the self-teacher network. \#channel of BiFPNc depends on the channel width parameter, $w$. Parameters and FLOPs are the ratio of those of the classifier network.}
		\label{table:bifpn_structure}
	\end{table}
	
	\begin{table*}[h]
		\centering
		\begin{tabular}{ccccccccc}
			\toprule
			Method & CIFAR-100 & TinyImageNet & CUB200 & MIT67 & Dogs & Stanford40 \\ \midrule
			Baseline & 73.80$\pm$\scriptsize{0.60} & 54.60$\pm$\scriptsize{0.33} & 51.72$\pm$\scriptsize{1.17} & 55.00$\pm$\scriptsize{0.97} & 63.38$\pm$\scriptsize{0.04} & 42.97$\pm$\scriptsize{0.66}\\
			FitNet & 76.65$\pm$\scriptsize{0.25} & 59.38$\pm$\scriptsize{0.10} & 58.97$\pm$\scriptsize{0.07} & 59.15$\pm$\scriptsize{0.41} & 67.18$\pm$\scriptsize{0.10} & 46.64$\pm$\scriptsize{0.24}\\
			ATT & \underline{77.16$\pm$\scriptsize{0.15}} & \textbf{59.83$\pm$\scriptsize{0.28}} &  \underline{59.21$\pm$\scriptsize{0.34}} &  \underline{59.33$\pm$\scriptsize{0.22}} &  \underline{67.54$\pm$\scriptsize{0.18}} & 47.04$\pm$\scriptsize{0.17}\\
			Overhaul & 74.59$\pm$\scriptsize{0.32} &  59.50$\pm$\scriptsize{0.09} & 58.82$\pm$\scriptsize{0.12} &  58.81$\pm$\scriptsize{0.58} & 66.43$\pm$\scriptsize{0.08} &  \underline{47.06$\pm$\scriptsize{0.26}}\\ 
			FRSKD & \textbf{77.71$\pm$\scriptsize{0.14}} & \underline{59.61$\pm$\scriptsize{0.31}} & \textbf{65.39$\pm$\scriptsize{0.13}} & \textbf{61.74$\pm$\scriptsize{0.67}} & \textbf{70.77$\pm$\scriptsize{0.20}} & \textbf{56.00$\pm$\scriptsize{1.19}}\\ \bottomrule
		\end{tabular}
		\caption{Performance comparison on knowledge distillation. ResNet18 is used as classifier network. The best performing model is indicated as boldface. The second-best model is indicated as underline.}
		\label{table:KD}
	\end{table*}
	
	\begin{table*}[h]
		\centering
		\begin{tabular}{ccccccccc}
			\toprule
			Method & CIFAR-100 & TinyImageNet & CUB200 & MIT67 & Dogs & Stanford40 \\ \midrule
			Baseline & 73.80$\pm$\scriptsize{0.60} & 54.60$\pm$\scriptsize{0.33} & 51.72$\pm$\scriptsize{1.17} & 55.00$\pm$\scriptsize{0.97} & 66.38$\pm$\scriptsize{0.04} & 42.97$\pm$\scriptsize{0.66}\\
			Mixup & 76.26$\pm$\scriptsize{0.41} & 56.28$\pm$\scriptsize{0.24} & 57.60$\pm$\scriptsize{0.42} & 56.77$\pm$\scriptsize{1.45} & 65.96$\pm$\scriptsize{0.03} & 47.15$\pm$\scriptsize{0.60}\\ 
			
			FRSKD + Mixup & 78.74$\pm$\scriptsize{0.19} & \underline{60.30$\pm$\scriptsize{0.38}} & \textbf{67.98$\pm$\scriptsize{0.58}} & \underline{62.11$\pm$\scriptsize{0.81}} & \underline{71.64$\pm$\scriptsize{0.29}} & \textbf{56.50$\pm$\scriptsize{0.36}}\\
			CutMix & \underline{79.23$\pm$\scriptsize{0.23}} & 58.97$\pm$\scriptsize{0.29} & 51.54$\pm$\scriptsize{1.12} & 60.87$\pm$\scriptsize{0.30} & 67.71$\pm$\scriptsize{0.14} & 46.90$\pm$\scriptsize{0.29}\\
			FRSKD + CutMix & \textbf{80.49$\pm$\scriptsize{0.05}} & \textbf{61.92$\pm$\scriptsize{0.11}} & \underline{65.92$\pm$\scriptsize{0.59}} & \textbf{66.19$\pm$\scriptsize{0.49}} & \textbf{72.81$\pm$\scriptsize{0.23}} & \underline{55.75$\pm$\scriptsize{0.43}}\\ \bottomrule
		\end{tabular}
		\caption{Performance of data augmentation method with FRSKD. ResNet18 is used as classifier network. The best performing model is indicated as boldface. The second-best model is indicated as underline.}
		\label{table:Data augmentation}
	\end{table*}
	
	\noindent\textbf{Ablation with the feature distillation methods}
	FRSKD is able to integrate diverse feature distillation methods, so we experiment such variations of the integrated feature distillations. To analyze the performance differences, we compare 1) exact feature distillation methods, FitNet and Overhaul distillation; and 2) attention transfer methods used in FRSKD. Table~\ref{table:feature_distill} shows that the attention transfer of FRSKD achieves better accuracy in various datasets than the accuracy from the integration of exact feature distillations. 
	
	\noindent\textbf{Structure of self-teacher network}
	In order to show the proposed self-teacher network efficiency, we experiment various self-teacher network structures, and Table~\ref{table:bifpn_structure} shows the experimented variations. BiFPN with high channel dimension (256) achieves the best performance, but the parameter and FLOPs of BiFPN are even larger or similar to the classifier network. In terms of efficiency, while BiFPN is an inadequate choice for the self-teacher network because of its large parameter size, BiFPNc with high channel dimension (256) shows a compatible performance to BiFPN with much less computations. Since the increased computation of BiFPNc is smaller than those of the classifier network, FRSKD is more efficient than the data augmentation based self-knowledge distillation methods, which use the classifier network in duplicates.
	
	\noindent\textbf{Compare to knowledge distillation}
	Knowledge distillation with a teacher network easily utilizes the refined feature-map and its soft label. Therefore, we compare the existing knowledge distillation and FRSKD, which play similar roles. Under the assumption that we have a pretrained teacher network, we compare exact feature distillation methods, FitNet and overhaul distillation~\cite{fitnets,overhaul}; and attention transfer methods~\cite{atts}. For knowledge distillation, we set the teacher network as the pretrained ResNet34 on each dataset, and we set the student network as an untrained ResNet18. For fair comparisons, each knowledge distillation method exploits the feature distillation as well as the soft label distillation. FRSKD utilizes ResNet18 as a classifier network to meet the identical conditions. Table~\ref{table:KD} shows that FRSKD outperforms the experimented knowledge distillation method with a pretrained teacher network on most datasets. 
	
	\noindent\textbf{Training with data augmentation}
	The method of data augmentation is compatible with FRSKD. To verify this compatibility of FRSKD, we experiment our proposed method with recent data augmentation methods. Mixup utilizes a convex combination between two images and their labels~\cite{mixup}. Cutmix mixes a pair of images and labels by cutting an image into patches and pasting a patch on the other image. It is known that such data augmentations improve the accuracy on the most datasets. Table~\ref{table:Data augmentation} shows that FRSKD has a large performance improvement when being used with data augmentations.
	
	\section{Conclusion}
	This paper presents a specialized neural network structure for self-knowledge distillation with top-down and bottom-up paths. The addition of these paths are expected to provide refined feature-maps and their soft labels to the classifier network. Moreover, the change of channel dimensions is applied to reduce the parameters while maintaining the feature-map refinement. Finally, FRSKD is able to apply self-knowledge distillation to the vision tasks of classification and semantic segmentation. We confirmed the large performance improvements quantitatively, and we show the efficiency of working mechanisms with various ablation studies.
	
	\noindent\small{\textbf{Acknowledgments} This research was supported by Basic Science Research Program through the National Research Foundation of Korea (NRF) funded by the Ministry of Education (NRF-2021R1A2C2009816).}
	
	%
	
	\section{Implementation Details}
	\noindent \textbf{Classification~~}
	WRN-16-2 consist of three blocks with channel dimensions of 32, 64, and 128 respectively, while ResNet18 consists of four blocks with channel dimensions of 64, 128, 256 and 512, respectively. We set the width hyperparameter, $w$, as two for all experiments on classification tasks in the paper. Therefore, channel dimensions of the self-teacher networks are 64, 128, 256 for WRN-16-2 and 128, 256, 512, 1028 for ResNet18. 
	
	For Mixup and Cutmix data augmentation, we need to set hyperparameter to model beta distribution that determine the mixing weights~\cite{cutmix,mixup}. For Mixup, we set hyperparameter as 0.2 for CIFAR-100 and TinyImageNet, and 0.3 for FGVR. For Cutmix, we set hyperparameter as 1.0 for all datasets.
	
	\noindent \textbf{Semantic segmentation~~}
	For semantic segmentation, we set width hyperparameter as one, and we set other settings for network architecture following \cite{bifpn}. We attach the self-teacher network with two repeated BiFPN layers and we do not change the channel dimension of BiFPN layers on semantic segmentation task. Therefore, classifier network with three BiFPN layer including backbone network and the self-teacher network are trained by cross entropy loss from ground truth label. Additionally, the self-teacher network perform distillation for classifier network by soft-label and features.
	
	Since the labeled dataset for semantic segmentation is insufficient, it is common to use a pretrained backbone network. We use pretrained efficientnet-b0 and efficientnet-b1 on ImageNet. Therefore, classifier network consists of pretrained backbone network and BiFPN layers. We apply warm-up and annealing technique to the hyperparameter of FRSKD, because distillation from the beginning of learning could rather be a hindrance to train the backbone network. We set warm-up epoch as 40 and after warm-up we adaptively increase the hyperparameters. We set hyperparameter $alpha$ as one and $beta$ as 50.
	
	\section{Sensitivity Analysis}
	\vspace{-1.5em}
	\begin{figure}[h]
		\centering
		\includegraphics[width=\columnwidth]{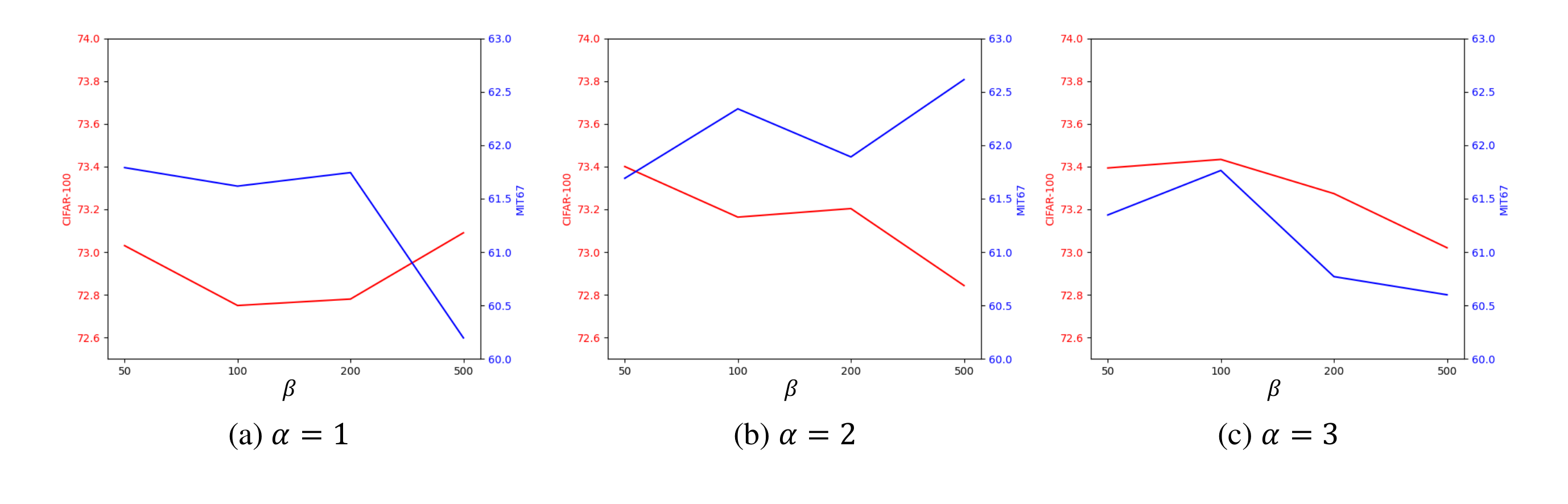}
		\caption{Sensitivity analysis for hyperparameter $\alpha$ and $\beta$. Red line indicates the accuracy of WRN-16-2 on CIFAR-100; and blue line indicates the accuracy of ResNet18 on MIT67. The accuracy is average of three repeated experiments.}
		\label{fig:sensitivity}
		\vspace{-1em}
	\end{figure}

	We evaluate FRSKD with varying hyperparameter values to investigate the effect of hyperparameters, $\alpha$ and $\beta$. We conduct experiments with $\alpha\in\{1,2,3\}$ and $\beta\in\{50,100,200,500\}$. Also, we set classifier network as WRN-16-2 on CIFAR-100 and ResNet18 on MIT67. Figure~\ref{fig:sensitivity} shows the accuracy of each dataset with varying hyperparameters. We keep all-settings except hyperparameters as same as Section 4.1. We find that FRSKD is robust on hyperparameter $\alpha$ and $\beta$, but different hyperparameters perform well for different datasets.
	
	\begin{figure*}[h]
		\centering
		\includegraphics[width=2\columnwidth]{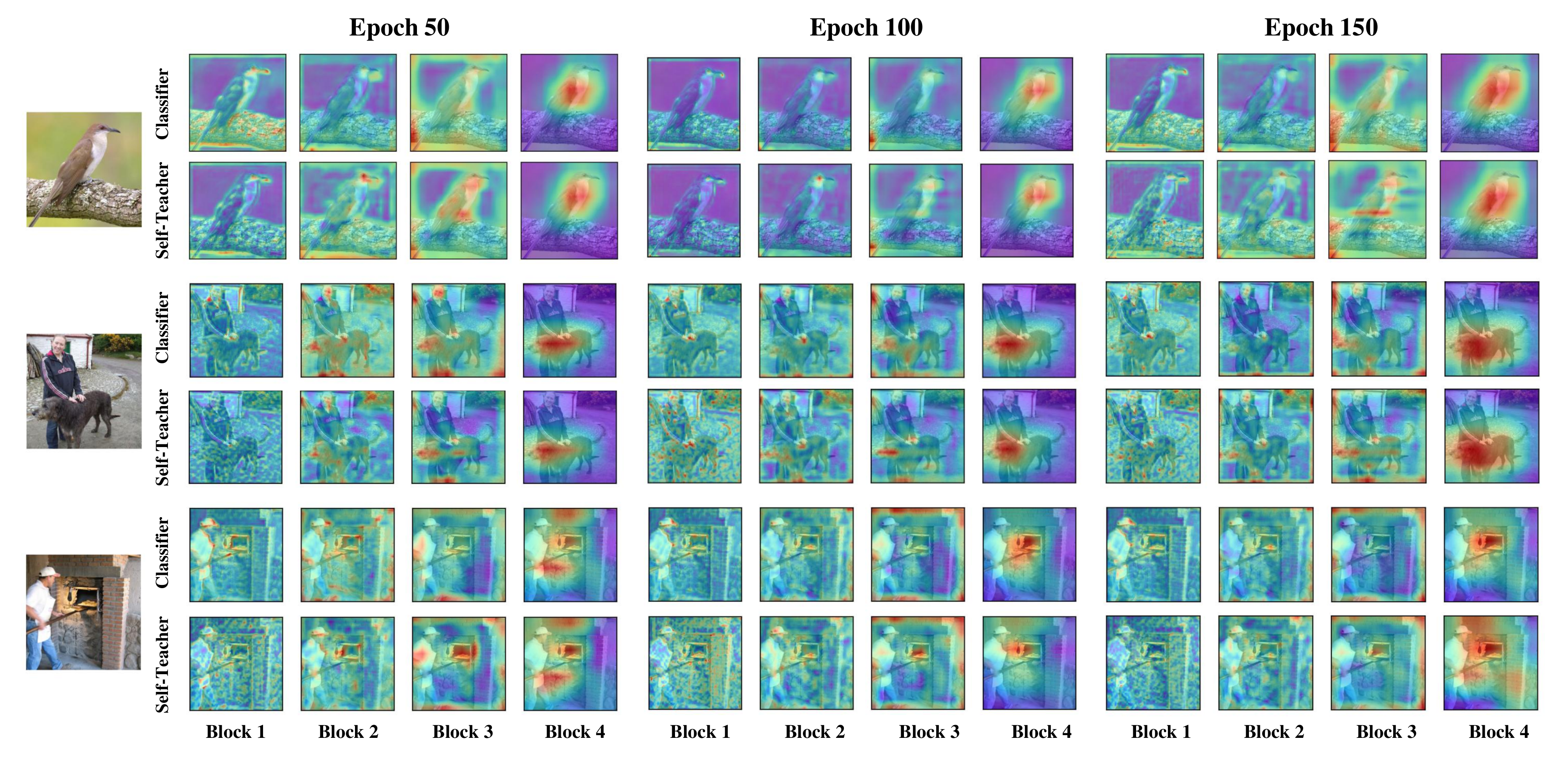}
		\caption{The block-wise attention map comparison between the classifier network and self-teacher network with varying epochs. From above each data is taken from CUB200, Dogs, and MIT67.}
		\label{fig:attention} 
		\vspace{-1.5em}
	\end{figure*}
	
	\section{Qualitative Attention Map Comparison}
	This section provides additional result of qualitative attention map comparison. We use a channel-wise pooling to the feature map as same as analysis of Section \textit{Further Analyses on FRSKD} in the paper. Figure~\ref{fig:attention} shows the attention map changes as the learning progresses. As the training progresses, Both the classifier network and the self-teacher network concentrate pm the main object. Additionally, the difference between concentration on the main object is bigger at the early epoch of the training than at the latter part of training.
	
	{\small
	\bibliographystyle{ieee_fullname}
	\bibliography{egbib}
	}

\end{document}